\documentclass{article}
\usepackage{spconf,amsmath,graphicx}
\usepackage{times}
\usepackage{latexsym}
\usepackage{multirow}
\usepackage{graphicx}
\usepackage{color}
\usepackage{amssymb}
\usepackage{subfigure}
\usepackage{adjustbox}
\usepackage{bm}
\usepackage{amsmath,multirow,booktabs,tabularx}
\usepackage{threeparttable}

\title{Noise-BERT: A Unified Perturbation-Robust Framework with Noise Alignment Pre-training for Noisy Slot Filling Task}
\name{
Jinxu Zhao\textsuperscript{*},
Guanting Dong\sthanks{The first two authors contribute equally.},
Yueyan Qiu,
Tingfeng Hui,
Xiaoshuai Song,
Daichi Guo,
Weiran Xu\sthanks{Weiran Xu is the corresponding author. Any questions please contact email: zhaojinxu@bupt.edu.cn }
}

\address{Beijing University of Posts and Telecommunications, China}

\begin{document}
%\ninept
%

\maketitle
\begin{abstract}
In a realistic dialogue system, the input information from users is often subject to various types of input perturbations, which affects the slot-filling task. Although rule-based data augmentation methods have achieved satisfactory results, they fail to exhibit the desired generalization when faced with unknown noise disturbances. In this study, we address the challenges posed by input perturbations in slot filling by proposing \textbf{Noise-BERT}, a unified Perturbation-Robust Framework with Noise Alignment Pre-training. Our framework incorporates two Noise Alignment Pre-training tasks: Slot Masked Prediction and Sentence Noisiness Discrimination, aiming to guide the pre-trained language model in capturing accurate slot information and noise distribution. During fine-tuning, we employ a contrastive learning loss to enhance the semantic representation of entities and labels. Additionally, we introduce an adversarial attack training strategy to improve the model's robustness. Experimental results demonstrate the superiority of our proposed approach over state-of-the-art models, and further analysis confirms its effectiveness and generalization ability.
\end{abstract}
\section{Introduction}
\label{sec:intro}

The slot filling task as s crucial in Task-Oriented Dialogue (TOD) systems \cite{wu2019transferable,hosseini2020simple,zeng2022semisupervised,lin2021leveraging,heck2020trippy,qixiang-etal-2022-exploiting,lee2021dialogue,zhao2022description,10193387,wu2023semantic,mekala2022zerotop} aims to extract specific information from the natural language input of users and fill it into predefined slots . Recently, several data-driven methods have achieved satisfactory performance in the slot filling and sequence labeling tasks \cite{zhao-etal-2022-entity,yan-etal-2020-adversarial,zhang-etal-2019-joint,wang-etal-2021-bridge,zhang-etal-2019-joint,liu2023robust,liu-lane-2016-joint,10095149,li2023generative,shi2023adaptive,10.1145/3583780.3614766,wang2023gptner}. However, these methods often underperform when applied to scenarios with perturbations. This is because the inconsistency between the training and test data distributions \cite{wu2021bridging}. The performance of models is significantly affected in the presence of diverse input disturbances, leading to a decline in their efficacy when confronted with noisy scenarios.

%--------------------------intro图-----------------------
\begin{figure}[t]
\centering
% \subfigure[Performance drop of baselines on Noise-SF\label{fig:intro1}]{
% \resizebox{.44\textwidth}{!}{\includegraphics{figures/intro_lstm.pdf}
% }
% }
\resizebox{.46\textwidth}{!}{\includegraphics{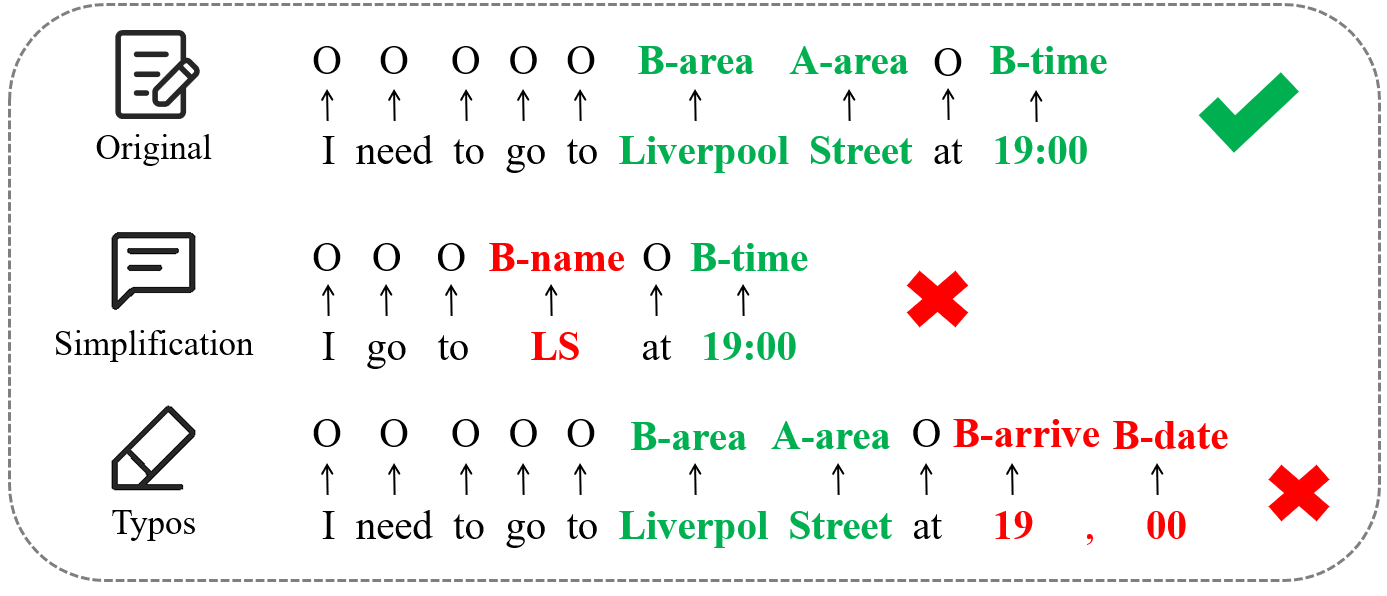}}
\vspace{-0.3cm} 
\caption{The impact of various types of input perturbations on the slot filling task in real dialogue systems}
\label{fig:intro} 
\vspace{-0.3cm} 
\end{figure}
%--------------------------intro图-----------------------

As shown in Figure \ref{fig:intro}, in real dialogue systems, deviations from the standard input format frequently occur due to the diversity of language expressions and input errors. For example, users may interact with the dialogue system in a format that does not comply with the prescribed format due to different expressive habits, yet their simplified expressions still convey the same intent. Additionally, upstream error-prone input systems, such as keyboard input errors or errors in speech recognition systems, can introduce interference to downstream models. Therefore, it is necessary to train a robust slot filling model to handle input perturbations.
%--------------------------main图-----------------------
\begin{figure*}[t]
\centering
% \subfigure[Performance drop of baselines on Noise-SF\label{fig:intro1}]{
% \resizebox{.44\textwidth}{!}{\includegraphics{figures/intro_lstm.pdf}
% }
% }
\resizebox{.95\textwidth}{!}{\includegraphics{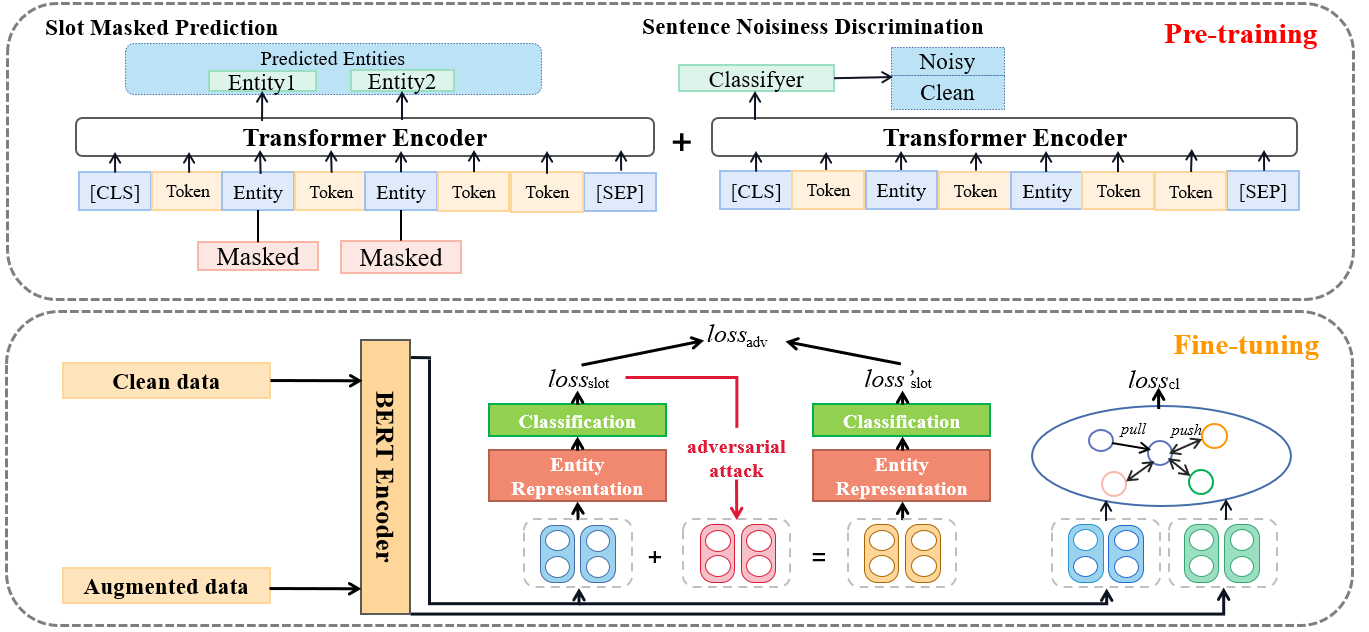}}

\caption{The overall architecture of our proposed Noise-BERT framework}
\label{fig:main} 

\end{figure*}
%--------------------------main图-----------------------

Most of the recent research focusing on the impact of input perturbations on Natural Language Processing (NLP) systems \cite{wu2021bridging,9747192,moradi2021evaluating,gui2021textflint,song-etal-2023-large,shah-etal-2019-robust} validates the robustness of models using noise datasets generated through fixed rules. However, these methods are designed for specific perturbations and can only be applied to specific types of noisy scenarios. To address the issue of unknown perturbations, PSSAT and CDMA \cite{dong2022pssat,10094766} employs a BART model to fit Twitter data and generate data that aligns with real-world scenarios for downstream models. Nevertheless, Twitter data often encompasses discussions on social media topics that harbor biases and deviations. Recently, A series of prompt-based generative models enhance the robustness of the model by introducing diverse demonstrations \cite{wu-etal-2022-incorporating,10.1007/978-3-031-44693-1_53,min2022rethinking,dong-etal-2023-demonsf,pan2023incontext,ouyang2022training,zhao-etal-2023-demosg} and instructions \cite{Luo2023ZeroShotSF,wang2022instructionner,xiao2024yayiuie,lei2023instructerc,wang2023instructuie}. As a consequence, the model's capacity for generalization in other domains or tasks may be compromised. Therefore, there is a need for more comprehensive research on noise disturbances to address various input perturbations.

To tackle these limitations, we propose Noise-BERT, a Unified Perturbation-Robust Framework with Noise Alignment Pre-training. Specifically, in the pre-training phase, we introduce two Noise Alignment Pre-training tasks, Slot Masked Prediction and Sentence Noisiness Discrimination, tailored to the characteristics of noisy slot filling data. These tasks provide guidance to the pre-trained language model in capturing accurate slot information and noise distribution in scenarios characterized by diverse input perturbations. In the fine-tuning phase, we employ contrastive learning loss to enhance the semantic representation of entities and labels and learn the semantic patterns of relevant slot entities. To improve the generalization ability of our model, we also propose an adversarial attack training strategy. This training strategy adds adversarial noise to the input in the direction of significantly increasing the model's classification loss. It significantly enhances the adaptability and robustness of our approach.

Our main contributions are threefold:
(1) We propose two Noise Alignment Pre-training tasks, tailored to the characteristics of noise data and slot filling data, aiming to provide guidance to the PLM in accurately capturing slot information and noise distribution.
(2) We introduce a multi-task framework that strengthens the representation of entity boundaries and label information through the application of contrastive learning loss. Simultaneously, we employ an adversarial attack training strategy to improve the robustness of the model.
(3) Experimental results demonstrate that our proposed framework outperforms state-of-the-art models. Further extensive analysis has provided additional evidence of its effectiveness and generalization capabilities.

\section{METHODOLOGY}

% \subsection{Overall Architecture}
% As illustrated in Figure \ref{fig:main}, during the pre-training phase, model learn slot information and noise distribution under noisy scenarios through two Noise Alignment Pre-training tasks: Slot Masked Prediction and Sentence Noisiness Discrimination.

% During the downstream training phase, we introduce a multitask framework aimed at strengthening the representation of entity boundary information and entity label pair information by applying contrastive learning loss. Furthermore, we propose an adversarial attack training strategy that adds adversarial noise to the input in the direction of increasing the model's classification loss. The joint optimization of these two tasks enhances the adaptability and robustness of our method.

\subsection{Multi-level Data Augmentation}
To enhance the generalization of the model across different noisy scenarios, we employ the data augmentation tool NLPAug \cite{ma2019nlpaug} to augment the clean training set. The data augmentation techniques used in this research can be classified into three categories as follows. Character-level augmentation: randomly add, delete and replace characters in one token. Word-level augmentation: randomly delete and insert words and replace words with homophones in one sentence. Sentence-level augmentation: replace sentences with synonymous ones.

% randomly add, delete and replace characters in one token with the probability $p$. 
% randomly delete and insert words and replace words with homophones in one sentence with probability $p$. 
% replace sentences with synonymous ones.

\subsection{Noise Alignment Pre-training}
In order to enable the model to learn contextual semantic information under various perturbations, we design the pre-training tasks from two perspectives. As shown in Figure \ref{fig:main}, from a fine-grained perspective, we introduce Slot Masked Prediction to allow the model to learn slot information. From a coarse-grained perspective, we introduce Sentence Noisiness Discrimination to enable the model to understand the distribution of noise. In the pre-training phase, the data utilized is identical to the data employed in the fine-tuning phase.

\textbf{Slot Masked Prediction (SMP) }: Given an input utterance $X =\{x_{1}, . . . , x_{N}\}$, we randomly replace $k$ entities in the input with the special \emph{[MASK]} symbol and aim to predict their original values. If an entity $e$ consists of multiple tokens, all of its component tokens are masked. Thus, the loss function can be formulated as follows:
\begin{equation}
\begin{aligned}
\mathcal{L}_{smp}=-{\sum}_{m=1}^{M}\log P(\bm{x_m})
\end{aligned}
\end{equation}

where $M$ is the total number of masked tokens and $P(\bm{x_m})$ is the predicted probability of the token $x_m$ over the vocabulary size.
%-----------main -----------
\begin{table*}[ht]
    \centering
    \renewcommand\arraystretch{1.1}
    \scalebox{0.89}{
        \begin{tabular}{l|c|c|c|c|c|c|c}
        \toprule
        \multirow{2}{*}{\textbf{Methods}} & \multirow{2}{*}{\textbf{Clean}} &\multicolumn{1}{c|}{\textbf{Character-level}} & \multicolumn{1}{c|}{\textbf{Word-level}}  & \multicolumn{3}{c|}{\textbf{Sentence-level}}& \multirow{2}{*}{\textbf{Overall}} \\
        \cline{3-7}
        \multicolumn{1}{l|}{} & \multicolumn{1}{c|}{} & \textbf{Typos} & \textbf{Speech} & \textbf{Paraphrase} & \textbf{Simplification} & \textbf{Verbose} & \multicolumn{1}{c}{} \\
        \hline
        Biaffine & 82.45 &58.47 &74.64 &72.43 &71.12 &69.48 &69.23 \\
        W2NER & 85.74 &61.55 &79.47 &83.71 &82.63 &77.55 &76.98 \\
        NAT($ \mathcal{L}_{aug} $) & 96.04 &64.73 &82.43 &87.70 &87.37 &82.86 &81.02 \\
        NAT($ \mathcal{L}_{stabil} $) & 96.02 &64.86 &82.71 &87.47 &87.36 &83.02  &81.08 \\
        PSSAT* \footnotemark{} & 95.83 &67.65 &83.67  &89.31 &86.79 &82.94 &82.07 \\
        \hline
        
        Noise-BERT  &\textbf{96.16($\pm$0.4)} & \textbf{71.15($\pm$1.2)} & \textbf{84.72($\pm$1.1)} & \textbf{90.27($\pm$0.5)} & \textbf{88.74($\pm$0.7)} & \textbf{84.89($\pm$0.6)} & \textbf{83.95($\pm$0.8)} \\ 
        
        -Pre-training  &91.37($\pm$0.5) &64.24($\pm$0.3) &79.45($\pm$0.2) &83.96($\pm$0.7)&84.21($\pm$0.3) &79.37($\pm$0.8) &78.25($\pm$0.5) \\
       
        \setlength{\parindent}{8ex} -SMP & 92.46($\pm$0.6) &64.75($\pm$0.9) &80.46($\pm$1.3) &85.73($\pm$1.2) &84.95($\pm$0.8) &80.21($\pm$0.3) &79.22($\pm$0.9)\\
        
        \setlength{\parindent}{8ex} -SND & 92.67($\pm$0.5) &65.42($\pm$1.4) &79.54($\pm$0.6) &84.27($\pm$0.4) &84.55($\pm$1.0)&79.68($\pm$0.8) &78.69($\pm$0.8) \\
        
        -Con & 93.16($\pm$0.6) &65.11($\pm$0.9) &81.26($\pm$0.8) &85.95($\pm$0.2) &85.43($\pm$0.9) &80.57($\pm$0.2) &79.66($\pm$0.6) \\
        
        -Adv & 91.52($\pm$0.3) &62.96($\pm$1.1)&80.37($\pm$1.2)&84.19($\pm$0.8)
        &83.74($\pm$0.5) &79.26($\pm$0.4)&78.10($\pm$0.8) \\
        \bottomrule
        \end{tabular}
    }
    \vspace{-0.1cm}
    \caption{The performance (F1 score) of the Noise-BERT on RADDLE. In Overall column, we calculate the average F1 of the five input perturbations. "-" denotes the model performance without a specific module.}
    \label{tab:main}
    \vspace{-0.4cm}
    
\end{table*}
\footnotetext{We have contacted the authors of PSSAT, and here are the results obtained through the code provided by them.}
%-----------main -----------

\textbf{Sentence Noisiness Discrimination (SND) }: Given an input utterance $X$ and its label $Y$, where $Y$ indicates whether the input $X$ is clean or noise data, we add a simple linear layer after the \emph{[CLS]} token to perform binary classification and determine whether the input sentence contains noise or not. The loss function can be formulated as follows: 
\begin{equation}
\begin{aligned}
\mathcal{L}_{snd} = -y \log(\hat{y}) - (1-y) \log(1-\hat{y})
\end{aligned}
\end{equation}

where $\hat{y}$ denote the true label, and $y$ represent the probability of the predicted label.

We sum the Slot Masked Prediction task loss and the Sentence Noisiness Discrimination task loss, and finally obtain the overall loss function $\mathcal L $: 
\begin{equation}
\begin{aligned}
\mathcal{L} = \alpha \mathcal L_{smp}+ (1-\alpha) \mathcal L_{snd}
\end{aligned}
\end{equation}
where $ L_{smp}$ and $L_{snd}$ denote the loss functions of the two
tasks. In our experiments, we set $\alpha = 0.6$.

\subsection{Noise Adaptation Fine-tune}
% Contrastive learning (CL) has achieved significant success in unsupervised visual representation learning \cite{tian2020contrastive,he2020momentum,misra2020self,chen2020simple} . 
\textbf{Contrastive Learning (CL) }Contrastive learning has achieved significant success in unsupervised visual representation learning\cite{tian2019contrastive,khosla2020supervised}. It is to learn representations by maximizing the consistency between different views of the same data examples. To advance the semantic representation of entities and labels under noisy scenarios, we introduce contrastive learning loss. Specifically, given the original input data, we obtain its hidden representations using BERT\cite{devlin2019bert} and then treat its corresponding augmented data as positive samples and other augmented data in the same batch as negative samples. Therefore, the contrastive learning loss is formulated as follows:
\begin{equation}
\begin{aligned}
\mathcal{L}_{cl} = -\log \frac{\exp(\text{sim}(q, \bm{k}^+)/\tau}{\sum_{i=0}^{N} \exp(\text{sim}(q, \bm{k}_i))/\tau}
\end{aligned}
\end{equation}

where $q$ denote the input data, $\bm {k}^+$, $\bm {k}_i$ represent positive and augmented samples in the same batch.

\textbf{Adversarial Attack Training (Adv) }In this section, we present an adversarial attack training strategy, as illustrated in Figure \ref{fig:main}, to enhance the robustness of the model. Given an input sentence, we obtain its hidden representations using BERT. Next, the model utilizes them to characterize the words and their relations in the sentence, extract candidate entity words, and classify them based on predicted probability distributions. Finally, we obtain the classification cross-entropy loss, denoted as $\mathcal L_{slot}$. Due to existing methods are often susceptible to noisy input, in addition to the classification entropy loss $\mathcal L_{slot}$, we apply the Fast Gradient Value (FGV) technique \cite{miyato2016adversarial,
vedula2020open} to approximate the worst-case perturbation as a noise vector: 
\begin{equation}
\begin{aligned}
{v}_{noise} = \epsilon \frac{g}{\|g\|} ;  \text{ where } g = \nabla_e \mathcal L_{slot}
\end{aligned}
\end{equation}

Here, the gradient represents the first-order derivative of the loss function $\mathcal L_{slot}$, and $e$ denotes the direction of rapid increase in the loss function. We perform normalization and use a small $\epsilon$  to ensure the approximation is reasonable. Then, we add the noise vector $v_{noise}$ and conduct a second forward pass, obtaining a new loss $\mathcal L'_{slot}$. Finally, we employ adversarial attack loss $\mathcal L_{adv}= \mathcal L_{slot}+ \mathcal L'_{slot}$. The adversarial noise enables the model to handle a significant amount of noisy input sentences, and experimental results demonstrate that the adversarial training strategy effectively improves the performance of our method.

We finally obtain the overall loss function $\mathcal L$: 
\begin{equation}
\begin{aligned}
\mathcal L = \beta \mathcal L_{cl}+ (1-\beta) \mathcal L_{adv}
\end{aligned}
\end{equation}

In our experiments, we set $\beta = 0.3$.

\section{Experiments}

\subsection{Dataset}
Based on RADDLE \cite{peng2020raddle} and SNIPS \cite{coucke2018snips}, we adhere to the evaluation set provided by PSSAT, which includes two settings: single perturbation and mixed perturbation. For the single perturbation setting, RADDLE is a crowd-sourced diagnostic evaluation dataset covering a broad range of real-world noisy texts for dialog systems. Typos are caused by non-standard abbreviations, while Speech arises from recognition and synthesis errors from ASR systems. Simplification refers to users using concise words to express their intentions, while Verbos erefers to users using redundant words to express the same intention. Paraphrase is also common among users who use different words or restate the text basedon their language habits. For mixed perturbation setting, based on SNIPS, we used textflint \cite{gui2021textflint} to introduce character-level perturbation Typos, Word-level perturbation Speech, sentence-level perturbation AppendIrr, and mixed perturbation to the test set and construct a multi-perturbation evaluation set.

\subsection{Results and Analysis }
\textbf{Main Results.} Table \ref{tab:main} presents the main results obtained on datasets with different levels of perturbations compared to other baselines. Our proposed method demonstrates a significant improvement over all previous methods when subjected to various types of input perturbations while maintaining the best performance on the clean test data.
Regarding fine-grained perturbations, our approach demonstrates a remarkable performance improvement of 5.2\% compared to the SOTA method PSSAT when subjected to Typos perturbation. For coarse-grained perturbations, Noise-BERT maintains strong performance, particularly with a 2.3\% enhancement under Verbose perturbations. These results provide strong evidence that Noise-BERT accurately captures slot mention information under fine-grained perturbations while simultaneously preserving the generalization capability for global semantic information under coarse-grained perturbations.

\textbf{Ablation Studies.} To examine the characteristics of the main components, we conduct an ablation study in Table \ref{tab:main}. We make the following observations:
1) The performance of Noise-BERT decreases when any component is removed, indicating that each designed part is necessary.
2) Compared to removing any individual task, eliminating the joint pretraining task results in a significant performance drop. This suggests that the joint pretraining objective plays a complementary role.
3) The impact of adversarial training is the most substantial when removed, indicating that adversarial training enables the learning of more robust feature representations and decision boundaries, thereby improving generalization to unknown samples.

%-----------mix noise -----------
\vspace{-0.3cm}
\begin{table}[ht]
    \centering
    \renewcommand\arraystretch{1.2}
    \scalebox{0.69}{
        \begin{tabular}{l|c|c|c|c}
        \toprule
        \multirow{2}*{\textbf{Methods}} & \textbf{Char+Word} & \textbf{Char+Sen} & \textbf{Word+Sent} & \textbf{Char+Word+Sen}\\
        \cline{2-5}
        \multicolumn{1}{l|}{} & \textbf{Typ.+Spe.} & \textbf{Typ.+App.} & \textbf{Spe.+App.} & \textbf{Typ.+Spe.+App.}\\
        \hline
        W2NER & 41.72 & 50.17 & 48.83 & 34.51 \\
        PSSAT & 44.55 & 54.34 & 51.86 & 36.67\\
        Noise-BERT & \textbf{48.86($\pm$0.5)} & \textbf{59.09($\pm$0.7)} & \textbf{54.42($\pm$0.2)} & \textbf{41.41($\pm$0.9)}  \\
        \bottomrule
        \end{tabular}
    }
    \vspace{-0.2cm}
    \caption{The performance (F1 score) of Noise-BERT under mixed perturbations on SNIPS.}
    \vspace{-0.3cm}
    \label{tab:mix}
\end{table}
%-----------mix noise -----------
\textbf{Mixed Perturbations Experiment.} In real dialogue scenarios, mixed perturbations often appear in one input utterance at the same time. To validate the robustness and generalization of our method in the face of more complex noise scenarios, we conducted a mixed perturbations experiment. As evidenced by the findings presented in Table \ref{tab:mix}, our approach showcases varying levels of improvement across three distinct categories of dual-level perturbations. Notably, when all three levels of noise are concurrently perturbed, the enhancements achieved by our approach are particularly prominent. These results indicate that Noise-BERT exhibits strong performance and robustness when faced with mixed perturbations, highlighting its effectiveness and stability in dealing with more challenging noise scenarios.

\begin{figure}[t]
    \vspace{-0.3cm}
    \centering
    \subfigure[PSSAT]{
        \includegraphics[width=.22\textwidth]{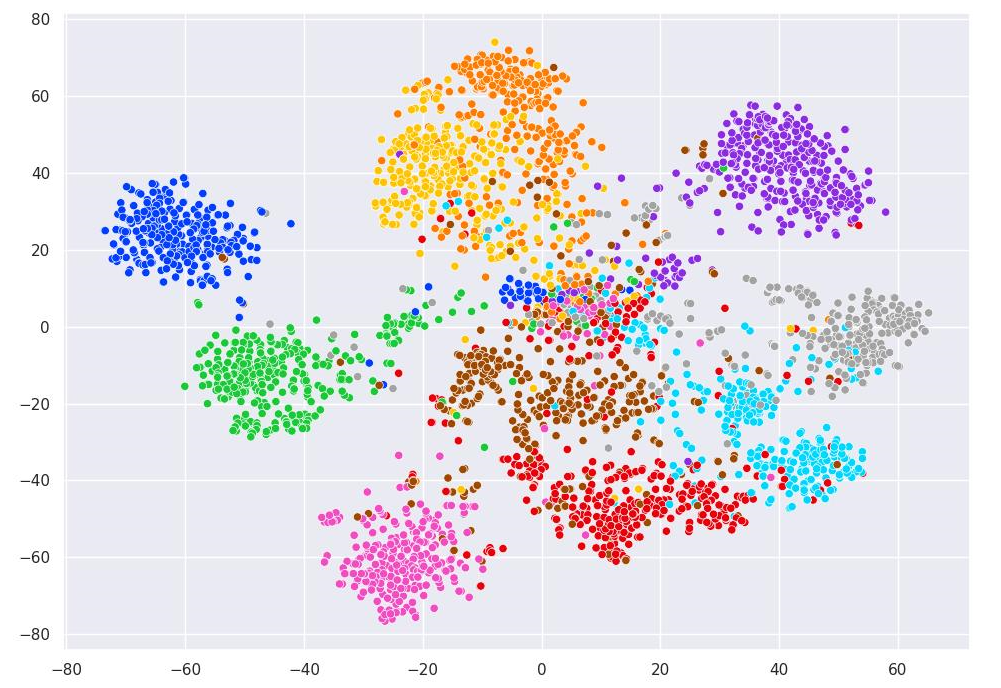}
        \label{noise}
    }
    \subfigure[Noise-BERT]{
	\includegraphics[width=.22\textwidth]{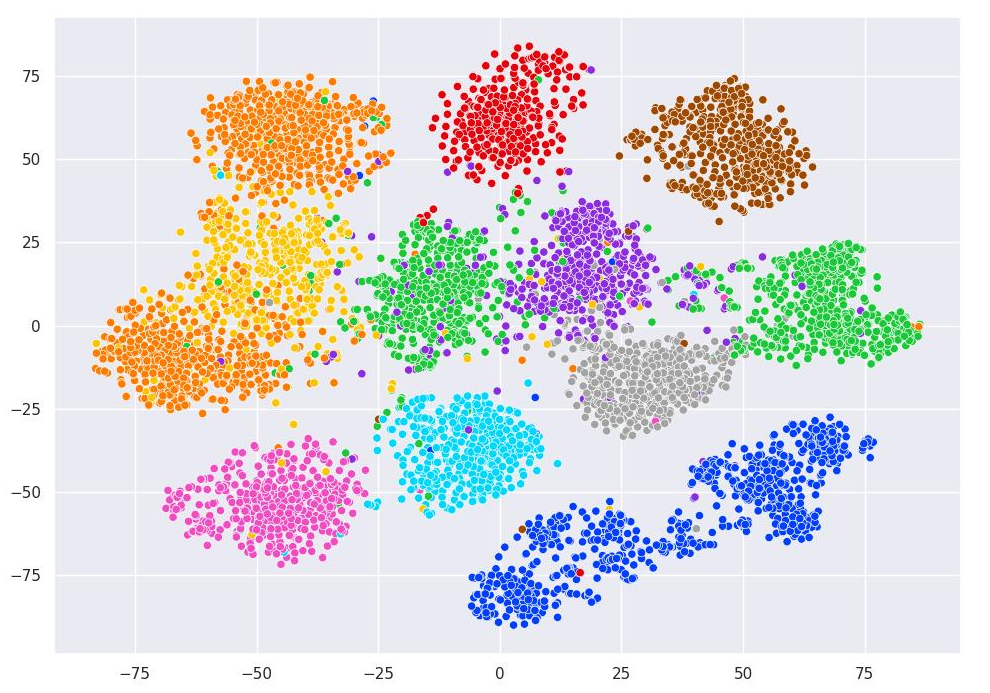}
        \label{noise2}
    }
    
    \caption{The t-SNE visualization of the entity representations with different types on mixed perturbation. } 
    \label{fig:visual}
    \vspace{-0.3cm}
\end{figure}

\textbf{Visualization.} We further conduct a detailed investigation to show how Noise-BERT learns the representations in the semantic space. We utilized data with mixed perturbation (Spe+App) from the SNIPS dataset and visualized the representations using the t-SNE \cite{van2008visualizing} toolkit. As shown in Figure \ref{fig:visual}, compared to the PSSAT, Noise-BERT increases the distances between representations of different categories and further compresses the distances within each category. The visualization demonstrates the capability of Noise-BERT to acquire meaningful and distinct representations within the semantic space. It offers further confirmation of its effectiveness in capturing semantic information related to entities across various categories, even in the presence of noise.

\section{Conclusion}
In conclusion, this study addresses the challenges associated with input perturbations in the slot-filling task within realistic dialogue systems.  We propose Noise-BERT, a Unified Perturbation-Robust Framework for Noisy Slot Filling Task. Specifically, We present two pre-training tasks, to capture accurate slot information and noise distribution in noisy scenarios. Furthermore, we incorporate a contrastive learning loss and employ an adversarial attack training strategy to enhance the adaptability and robustness of our approach during the fine-tuning stage. Experimental results demonstrate that Noise-BERT outperforms the previous SOTA methods in terms of overall performance. Extensive analysis further validates the effectiveness and generalization of our approach.

% References should be produced using the bibtex program from suitable
% BiBTeX files (here: strings, refs, manuals). The IEEEbib.bst bibliography
% style file from IEEE produces unsorted bibliography list.
% -------------------------------------------------------------------------
\bibliographystyle{IEEEbib}
\bibliography{Template}

\end{document}